\documentclass[runningheads]{llncs}
\usepackage[T1]{fontenc}
\usepackage{graphicx,verbatim}
\usepackage{amsmath}
\usepackage{amsfonts}
\usepackage{booktabs}
\begin{document}
\title{Gaussian Spatial Priors for Anatomy-Aware Object Detection in Surgical Videos}
%
\author{Yunfan Li\inst{1} \and
Artem Shmelev\inst{2} \and
Himanshu Gupta\inst{1}}
\authorrunning{Y. Li et al.}
\institute{Stony Brook University, Stony Brook NY 11794, USA \\
\email{yunfan.li@stonybrook.edu} \and
Stony Brook University Hospital, Stony Brook NY 11794, USA}
  
\maketitle              
\begin{abstract}
Detecting anatomical structures in surgical video is essential for intraoperative safety frameworks such as the Critical View of Myopectineal Orifice (CVMPO) in inguinal hernia repair. While prominent structures like the Cooper's Ligament and Triangle of Doom are reliably detected by standard methods, smaller structures such as the epigastric vessels remain challenging due to their visual ambiguity and intermittent visibility. We observe that the spatial relationship between structures is anatomically constrained, and propose a Gaussian Spatial Prior (GSP) module that encodes this relationship as a compact, parametric bias injected into the self-attention of a DAB-DETR decoder. The prior is computed offline from training annotations as a small set of frozen Gaussian parameters and recomputed at each decoder layer using the iteratively refined reference points. On a dataset of inguinal hernia repair videos with 5-fold cross-validation, GSP improves dependent class detection by $+33.5\%$ ($\text{AP}_{50}$) over DAB-DETR and $+53.9\%$ over YOLOv26, while also improving anchor detection by $+6.0\%$. These gains are statistically significant across all folds ($p=0.012$, paired $t-$test).


\end{abstract}
\section{Introduction}

Inguinal hernia repair is one of the most common surgical procedures, with approximately 800,000 operations performed annually in the United States~\cite{five-year}. Despite well-documented advantages of minimally invasive surgery (MIS)---including faster recovery, reduced pain and wound complications, and the ability to diagnose bilateral and occult hernias through a single posterior mesh reinforcement~\cite{herniasurge}—the MIS approach accounts for only a minority of repairs, with open mesh repair (Lichtenstein-Amid) remaining the dominant technique~\cite{five-year}. The MIS approach carries a significant learning curve, estimated at 50–100 procedures for laparoscopic repair~\cite{herniasurge,sivakumar} and 20–35 for robotic-assisted repair~\cite{solani}. Surgeons early in this curve experience elevated rates of complications and recurrence, frequently attributable to insufficient mesh overlap of the myopectineal orifice (MPO), inadequate parietalization of cord structures, and missed anatomical defects~\cite{felix,sato}.

To mitigate these failure modes, Daes and Felix~\cite{daes} proposed the Critical View of the Myopectineal Orifice (CVMPO), a standardized framework encompassing the identification of key anatomical landmarks and the completion of essential dissection steps prior to mesh placement. Subsequent work by Claus et al.~\cite{claus} and Marmolejo et al.~\cite{marmolejo} augmented CVMPO with scoring systems anchored to five triangular zones delineated by an inverted-Y configuration of anatomical structures. Analogous to the Critical View of Safety (CVS) in laparoscopic cholecystectomy~\cite{strasberg}, CVMPO aims to reduce adverse events by enforcing systematic anatomical verification. Yet unlike CVS, which has received substantial attention from the computer vision community~\cite{madani,murali,li,mascagni}, automated CVMPO assessment remains largely unexplored.

The only prior work addressing AI-based CVMPO assessment is that of Takeuchi et al.~\cite{takuchi}, who applied a standard DETR~\cite{detr} model to identify anatomical landmarks from the first seven CVMPO steps, achieving a mean average precision (mAP) of 51.2\% at the frame level. Their approach applies off-the-shelf detectors without incorporating the structured spatial relationships between landmarks, leaving considerable room for improvement—particularly in a safety-critical surgical context.

A fundamental challenge in surgical anatomy detection is that many anatomical structures are small, lack distinctive visual features, and are intermittently visible due to occlusion by tissue or instruments. Standard object detectors treat each class independently, relying on content-driven attention to learn inter-object relationships entirely from data. When training data is scarce—as is inherent to rare surgical conditions and novel clinical formulations—this purely data-driven approach is insufficient. However, the spatial layout of posterior groin anatomy is well characterized: given the positions of consistently visible structures such as Cooper's Ligament (CL) and the Triangle of Doom (DOOM), the expected locations of less prominent landmarks follow predictable, anatomically constrained distributions.

In this work, we propose to encode this domain knowledge as a parametric spatial prior and inject it directly into the detection architecture. Specifically, we augment DAB-DETR~\cite{dab-detr}, a transformer-based detector that uses explicit 4D anchor box coordinates as queries, with a Gaussian Spatial Prior (GSP) module. The GSP encodes compact Gaussian distributions over normalized spatial offsets between anatomical class pairs from training annotations, and biases the decoder's self-attention toward spatially plausible query configurations at every refinement layer. 

We evaluate our approach on a dataset of robotic transabdominal preperitoneal (TAPP) inguinal hernia repair videos annotated with bounding boxes for 3 anatomical structures. To our knowledge, this is the first work to (1) introduce the problem of automated anatomical landmark detection for CVMPO assessment to the computer vision community, and (2) propose a parametric spatial prior mechanism for Transformer-based surgical anatomy detection. Our experiments demonstrate that the GSP substantially improves detection performance over the standard DAB-DETR baseline, with particular gains on the dependent anatomical classes whose detection benefits most from spatial context.

\section{Methods}
Our method augments a transformer-based object detector with a structured spatial prior derived from the anatomical layout of surgical scenes. We first describe the partitoining of anatomical classes into anchors and dependents (Section~\ref{sec21}), and then review the base DAB-DETR architecture whose explicit anchor box queries provide the spatial interface for our prior (Section~\ref{sec22}). Section~\ref{sec23} introduces the Gaussian Spatial Prior (GSP) module, detailing the offline computation of prior distributions, the evaluation of pairwise spatial compatibility at inference, and the injection of the resulting bias into the decoder's self-attention. Finally, we summarize the training procedure in Section~\ref{sec24}. 

\subsection{Anchor-Dependent Class Structure \label{sec21}}

We focus on anatomical structures relevant to the Critical View of Myopectineal Orifice (CVMPO) safety framework for inguinal hernia repair. Of those, Cooper's Ligaments (CL) and Triangle of Doom (DOOM) are two important structures for surgeons to understand the scene and assess the progress of dissection. Moreover, they are consistently detectable and visually prominent across frames. We designate these as anchor classes $\mathcal{A}=\{a_1,a_2\}$. The remaining eight classes, which tend to be less distinct and intermittently visible, constitute the dependent classes $\mathcal{D}$.

This partitioning reflects a structural property of the inguinal hernia repair scene: given the positions of anchor structures, the locations of dependent structures follow predictable spatial distributions. We encode these distributions as parametric Gaussian priors and inject them as an inductive bias into the detector.

\subsection{Base DAB-DETR Architecture \label{sec22}}
We build on DAB-DETR~\cite{dab-detr}, which directly uses 4D box coordinates $\mathbf{a}_q=(x_q,y_q,w_q,h_q)$ as queries in the Transformer decoder and dynamically updates them layer by layer. Each query comprises a dual representation: a positional query (the anchor box) and a content query (a decoder embedding). Using box coordinates as positional queries provides explicit positional priors that improve query-to-feature similarity, while also enabling modulation of the positional attention map using the anchor width and height. This formulation is central to our approach, as it exposes interpretable spatial coordinates to which anatomical priors can be directly applied.
\subsubsection{Encoder.} A ResNet-50 backbone extracts features from the input frame. The flattened feature map is processed by a stack of self-attention layers with learned positional encoding.
\subsubsection{Decoder.} Each decoder layer applies self-attention among object queries and cross-attention to encoder memory. Query positions are derived from sinusoidal embeddings of the current anchor box coordinates, and the width and height components modulate the cross-attention map to achieve scale-aware feature extraction. The anchor boxes are updated layer by layer through residual refinements predicted by a shared bounding box head, progressively approaching the target objects.

\subsection{Gaussian Spatial Prior \label{sec23}}
Self-attention in DETR-based decoders is purely content-driven: inter-object spatial relations must be learned entirely from data. In surgical anatomy detection, where annotated data is scarce and spatial arrangement of structure is anatomically constrained, this imposes an unnecessary learning burden. We address this with a Gaussian Spatial Prior (GSP) module that biases decoder self-attention toward spatially plausible query configurations. The GSP operates in three stages: offline prior computation, pairwise spatial compatibility scoring, and attention bias injection.
\subsubsection{Offline Prior Computation.}
From training annotations, we compute parametric Gaussian distributions over the relative spatial offset between each dependent-anchor class pair. For each pair $(d_m,a_n)$ with $d_m \in \mathcal{D}, a_n \in \mathcal{A}$, we collect co-occurring instances across training data and compute normalized offsets:
\begin{equation}
    \delta x_{mn}=\frac{x^{(d)}-x^{(a)}}{w^{(a)}}, \delta y_{mn} = \frac{y^{(d)}-y^{(a)}}{h^{(a)}}
\end{equation}
where $(x^{(d)},y^{(d)})$ denotes the center coordinate of a dependent instance and $(x^{(a)}, y^{(a)}, w^{(a)}, h^{(a)})$ denotes the center and dimensions of an anchor instance. Normalization by anchor box size renders the offsets scale-invariant. We then fit a 2D axis-aligned Gaussian per class pair:
\begin{equation}
    \mu_{mn}=(\overline{\delta x_{mn}}, \overline{\delta y_{mn}}), \sigma_{mn} = (\text{std}(\delta x_{mn}), \text{std}(\delta y_{mn}))
\end{equation}

This yields $|\mathcal{D}|\times|\mathcal{A}|$ Gaussians, each described by four scalars $(\mu_x, \mu_y, \sigma_x, \sigma_y)$. Standard deviations are clamped to a minimum of 0.1 to prevent degenerate distributions. These parameters are pre-computed and stored as fixed buffers, and not updated during training.

\subsubsection{Pairwise Spatial Compatibility.}

At each decoder layer, given current reference points $\{\mathbf{r}_i\}_{i=1}^{N_q}$ for all $N_q$ queries, we compute spatial compatibility for every query pair $(i,j)$. The normalized offset from query $j$ to query $i$ is:
\begin{equation}
    \Delta x_{ij} = \frac{x_i-x_j}{w_j}, \Delta y_{ij} = \frac{y_i-y_j}{h_j}
\end{equation}
Each offset $\Delta_{ij} = (\Delta x_{ij}, \Delta y_{ij})$ is evaluated against all $|\mathcal{D}|\times|\mathcal{A}|$ Gaussians. The score of the $k$-th distribution is:
\begin{equation}
    g_k(\Delta_{ij}) = \text{exp}(-\frac{1}{2}\sum_{d \in \{x,y\}}(\frac{\Delta d_{ij}-\mu_k^d}{\sigma_k^d})^2)
\end{equation}
Scores are aggregated via a class-agnostic maximum over all distributions:
\begin{equation}
    s_{ij}=\max_{k=1,\cdots,|\mathcal{D}|\times|\mathcal{A}|} g_k(\Delta_{ij})
\end{equation}
This selects the best-matching anatomical relationship for each query pair without requiring class prediction at this stage.
\subsubsection{Attention Bias Injection.} 
The compatibility scores are converted to an additive attention bias. We first zero-center each row: 
\begin{equation}
    \hat{s}_{ij}=s_{ij}-\frac{1}{N_q}\sum_{j'=1}^{N_q}s_{ij'}
\end{equation}
Since softmax is shift-invariant, a uniform bias has no effect on attention weights; zero-centering ensures that only relative spatial compatibility influences the distribution. Offsets consistent with known anatomical relationships receive positive bias, while implausible offsets receive negative bias.

The centered bias is scaled by a hyperparameter $\lambda$ and shared across all attention heads:
\begin{equation}
    \mathbf{B}=\lambda \cdot \hat{\mathbf{S}} \in \mathbb{R}^{N_q \times N_q}
\end{equation}
$\mathbf{B}$ is added to the self-attention logits prior to softmax:
\begin{equation}
    \text{Attention}(\mathbf{Q}, \mathbf{K}, \mathbf{V})=\text{softmax}(\frac{\mathbf{Q}\mathbf{K}^{\top}}{\sqrt{d_k}}+\mathbf{B})\mathbf{V}
\end{equation}

Because DAB-DETR iteratively refines reference points, the bias is recomputed at every decoder layer using the updated positions, allowing the prior to progressively sharpen as predictions converge.


\subsection{Training \label{sec24}}
We train the model end-to-end using the standard DETR bipartite matching loss, comprising focal loss for classification, L1 loss for box regression, and generalized IoU loss. The Gaussian prior parameters remain fixed while only the base DAB-DETR parameters are optimized. We use 5-fold cross-validation partitioned at the video level to prevent patient overlap between training and validation splits.
\section{Experiments}
\subsection{Dataset and Evaluation Protocol}
We evaluate on a proprietary dataset of laparoscopic inguinal hernia repair videos annotated with bounding boxes for 3 key anatomical structures for CVMPO assessment: Cooper's Ligaments (CL), Triangle of Doom (DOOM), and Epigastric Vessels (EPI). The choice of these 3 classes is non-trivial: the relative positions of the three establish the orientation of the entire surgical scene, enabling surgeons to locate other visible or latent anatomical structures.

In total, we selected 619 frames from 119 IHR videos, and performed 5-fold cross validation partitioned at the video level to prevent patient-level data leakage. We use standard COCO-style Average Precision at IoU thresholds 0.50:0.95 (AP) and 0.50 ($\text{AP}_{50}$), reported both per class and as the mean across classes ($\text{mAP}$, $\text{mAP}_{50}$).

\subsection{Implementation Details}
All DETR-based models use a ResNet-50 backbone pretrained on COCO 2017 dataset. We train for 50 epochs using AdamW with a learning rate of $1e^{-4}$ and weight decay of $1e^{-4}$. The learning rate is reduced by a factor of 10 at epoch 40. The Gaussian prior parameters are computed from training annotations in each fold and stored as frozen buffers. All other training settings are default per original methods. We compare our method with both CNN-based and Transformer-based baselines, and we vary the prior strength $\lambda \in \{1,3,5,7\}$. 

\subsection{Results}
Table~\ref{tab:mean} and Table~\ref{tab:perclass} present the main detection results. We report per-class $\text{AP}$ and $\text{AP}_{50}$ alongside mean metrics to distinguish effects on anchor versus dependent classes. We note that for IHR scenes, $\text{AP}_{50}$ is more clinically meaningful than strict $\text{mAP}$: the goal is to reliably localize structures to guide the surgeons' attention, not to produce pixel-precise boundaries. Furthermore, unlike general objects, anatomical structures like ligaments and vessels are non-rigid and visually homogeneous, where precise bounding boxes often cannot be determined.
All baselines exhibit the same pattern: adequate anchor detection but poor EPI performance. YOLOv26 has the highest mAP, mainly due to strong performance on anchor classes, while having the lowest $\text{AP}_{50}$ on EPI. This confirms that strong overall detection does not address the unique challenge of dependent class in IHR scenes. 

While DAB-DETR underperforms vanilla DETR, the introduction of GSP consistently improves upon DAB-DETR, achieving best $\text{mAP}^{50}$ result at $\lambda = 3$, a $+9.3\%$ relative improvement over DAB-DETR and $+9.5\%$ over YOLOv26. The improvement is most pronounced on the dependent class: EPI $\text{AP}_{50}$ rises to 0.283, surpassing DAB-DETR by $+33.5\%$ and YOLOv26 by $+53.9\%$. Crucially, the prior also benefits anchor detection --- combined anchor $\text{AP}_{50}$ rises from 0.784 (DAB-DETR) to 0.831 ($+6.0\%$), indicating that the spatial prior regularised the full query set toward spatially cohherent configurations. GSP outperforms DAB-DETR on $\text{mAP}$ and $\text{mAP}_{50}$ in all five folds, and paired $t-$tests confirm significance ($p=0.012$ for both; one-sided). The prior also reduces cross-fold variability for EPI, stabilizing detection across patient-level splits.

\begin{table}[t]
\centering
\caption{Mean detection performance (mean $\pm$ std over 5 folds). $^\dagger$Paired $t$-test vs.\ DAB-DETR, $p < 0.05$ (one-sided). \textbf{Bold} = best; \underline{underlined} = second best.}
\label{tab:mean}
\small
\begin{tabular}{l cc}
\toprule
Method & $\text{mAP}$ & $\text{mAP}_{50}$ \\
\midrule
YOLOv26 & \textbf{.262}{\scriptsize$\pm$.019} & .592{\scriptsize$\pm$.023} \\
DETR & .253{\scriptsize$\pm$.023} & .615{\scriptsize$\pm$.044} \\
DAB-DETR & .238{\scriptsize$\pm$.030} & .593{\scriptsize$\pm$.057} \\
\midrule
GSP ($\lambda\!=\!3$) & \underline{.256}{\scriptsize$\pm$.021}$^\dagger$ & \textbf{.648}{\scriptsize$\pm$.034}$^\dagger$ \\
\bottomrule
\end{tabular}
\end{table}

\begin{table}[t]
\centering
\caption{Per-class detection performance.}
\label{tab:perclass}
\small
\setlength{\tabcolsep}{3.5pt}
\begin{tabular}{l cc cc cc}
\toprule
& \multicolumn{2}{c}{CL (anchor)} & \multicolumn{2}{c}{DOOM (anchor)} & \multicolumn{2}{c}{EPI (dependent)} \\
\cmidrule(lr){2-3} \cmidrule(lr){4-5} \cmidrule(lr){6-7}
Method & $\text{AP}$ & $\text{AP}_{50}$ & $\text{AP}$ & $\text{AP}_{50}$ & $\text{AP}$ & $\text{AP}_{50}$ \\
\midrule
YOLOv26 & \underline{.335}{\scriptsize$\pm$.044} & .793{\scriptsize$\pm$.070} & \textbf{.383}{\scriptsize$\pm$.047} & .800{\scriptsize$\pm$.070} & .068{\scriptsize$\pm$.026} & .184{\scriptsize$\pm$.059} \\
DETR (50ep) & \textbf{.340}{\scriptsize$\pm$.019} & .829{\scriptsize$\pm$.053} & \underline{.352}{\scriptsize$\pm$.038} & .796{\scriptsize$\pm$.061} & .066{\scriptsize$\pm$.033} & .222{\scriptsize$\pm$.104} \\
DAB-DETR & .316{\scriptsize$\pm$.039} & .802{\scriptsize$\pm$.047} & .331{\scriptsize$\pm$.042} & .766{\scriptsize$\pm$.084} & .067{\scriptsize$\pm$.034} & .212{\scriptsize$\pm$.095} \\
\midrule
GSP ($\lambda\!=\!3$) & .342{\scriptsize$\pm$.037} & \textbf{.835}{\scriptsize$\pm$.054} & .349{\scriptsize$\pm$.026} & \textbf{.827}{\scriptsize$\pm$.056} & \textbf{.078}{\scriptsize$\pm$.020} & \textbf{.283}{\scriptsize$\pm$.070} \\
\bottomrule
\end{tabular}
\end{table}

\subsection{Ablation: Prior Strength}
Table~\ref{tab:ablation} illustrates the effect of prior strength $\lambda$ on detection performance. Anchor detection is stable across all $\lambda$, confirming that the prior does not disturb the detection of anchors, while modulating dependent class detection. Performance peaks at $\lambda=3$ on most metrics, which we select for comparison.

\begin{table}[t]
\centering
\caption{Effect of prior strength $\lambda$ (mean over 5 folds). Anchor $\text{AP}_{50}$ is the average of CL and DOOM $\text{AP}_{50}$. \textbf{Bold} = best per column.}
\label{tab:ablation}
\small
\begin{tabular}{c ccccc}
\toprule
$\lambda$ & Anch.\ $\text{AP}_{50}$ & EPI AP & EPI $\text{AP}_{50}$ & mAP & $\text{mAP}_{50}$ \\
\midrule
1 & .821 & \textbf{.081} & .257 & .255 & .633 \\
3 & \textbf{.831} & .078 & \textbf{.283} & \textbf{.256} & \textbf{.648} \\
5 & .821 & .059 & .180 & .248 & .607 \\
7 & .816 & .074 & .257 & .251 & .630 \\
\bottomrule
\end{tabular}
\end{table}

\section{Conclusion}

We introduced the Gaussian Spatial Prior, a parameter-free module that injects anatomical spatial knowledge into the self-attention of a DAB-DETR decoder. By encoding the spatial relationship between anchor and dependent structures as a compact set of Gaussian distributions, the GSP biases query interactions toward anatomically plausible configurations without modifying the detector architecture or adding learned parameters. On laparoscopic inguinal hernia repair videos, this improves detection of the challenging epigastric vessels by +33.5\% AP50 over DAB-DETR and +53.9\% over YOLOv26, with gains that are statistically significant and consistent across all cross-validation folds.
Future work will extend the approach to additional dependent classes and procedures, explore graph-based modeling of higher-order spatial relationships, and integrate detection with downstream CVPMO criteria assessment toward a complete intraoperative safety system.

%
%
%
%

\end{document}